\documentclass[letterpaper]{article} % DO NOT CHANGE THIS
\usepackage{aaai25}
\usepackage[table,xcdraw, dvipsnames]{xcolor}
\usepackage{times}  % DO NOT CHANGE THIS
\usepackage{helvet}  % DO NOT CHANGE THIS
\usepackage{courier}  % DO NOT CHANGE THIS
\usepackage[hyphens]{url}  % DO NOT CHANGE THIS
\usepackage{graphicx} % DO NOT CHANGE THIS
\usepackage{natbib}  % DO NOT CHANGE THIS AND DO NOT ADD ANY OPTIONS TO IT
\usepackage{caption} % DO NOT CHANGE THIS AND DO NOT ADD ANY OPTIONS TO IT
\usepackage{algorithm}
\usepackage{algorithmic}
\usepackage{amsfonts}
\usepackage{amsmath}
\usepackage{bm}  % DO NOT CHANGE THIS

\usepackage[color=red]{todonotes}

\usepackage{newfloat}
\usepackage{listings}
\DeclareCaptionStyle{ruled}{labelfont=normalfont,labelsep=colon,strut=off} % DO NOT CHANGE THIS
\floatstyle{ruled}
\newfloat{listing}{tb}{lst}{}
\floatname{listing}{Listing}
\usepackage[utf8]{inputenc} % allow utf-8 input
\usepackage{url}            % simple URL typesetting
\usepackage{booktabs}       % professional-quality tables
\usepackage{amsfonts}       % blackboard math symbols
\usepackage{nicefrac}       % compact symbols for 1/2, etc.
\usepackage{microtype}      % microtypography

\usepackage{diagbox} % Include this package
\usepackage{amssymb} % For additional symbols
\usepackage{booktabs}
\usepackage{multirow}
\usepackage{graphicx} 
\usepackage{subcaption}
\usepackage{amsmath}

\definecolor{darkgreen}{HTML}{409120} % RGB color code
\definecolor{darkred}{HTML}{c90a0a}
\usepackage{ntheorem}
\theoremseparator{:}

\def\showauthors@on{T}

\title{The Master Key Filters Hypothesis:\\ Deep Filters Are General}
\author {
    Zahra Babaiee\textsuperscript{\rm 1}\equalcontrib,
    Peyman M. Kiassari\textsuperscript{\rm 1}\equalcontrib,
    Daniela Rus\textsuperscript{\rm 2},
    Radu Grosu\textsuperscript{\rm 1}
}
\affiliations {
    \textsuperscript{\rm 1}Technische Universit{\"a}t Wien, \textsuperscript{\rm 2}Massachusetts Institute of Technology\\
    zahra.babaiee@tuwien.ac.at, 
    peyman.kiasari@tuwien.ac.at, 
    rus@mit.edu,
    radu.grosu@tuwien.ac.at
}

\usepackage{bibentry}
\begin{document}

\maketitle

\begin{abstract}
This paper challenges the prevailing view that convolutional neural network (CNN) filters become increasingly specialized in deeper layers. Motivated by recent observations of clusterable repeating patterns in depthwise separable CNNs (DS-CNNs) trained on ImageNet, we extend this investigation across various domains and datasets. Our analysis of DS-CNNs reveals that deep filters maintain generality, contradicting the expected transition to class-specific filters. We demonstrate the generalizability of these filters through transfer learning experiments, showing that frozen filters from models trained on different datasets perform well and can be further improved when sourced from larger datasets. Our findings indicate that spatial features learned by depthwise separable convolutions remain generic across all layers, domains, and architectures. This research provides new insights into the nature of generalization in neural networks, particularly in DS-CNNs, and has significant implications for transfer learning and model design.

%Generalization in neural networks is commonly evaluated based on a model's performance on unseen test samples or its ability to adapt to new domains. In this paper, we investigate generalization by examining the learned weights of depthwise separable convolutional neural networks. We observe that models trained on different datasets exhibit similar general patterns in their depthwise convolutional filters, regardless of the dataset domain and model architecture, suggesting that these networks learn a general function across various domains and model architectures. To further support this claim, we demonstrate the transferability of these filters among different domains. Notably, networks with frozen filters from a model trained on a different dataset tend to perform well, and their performance can be enhanced if the source model was trained on a larger domain and exhibited better performance. Furthermore, we challenge the notion of specific feature learning in deep layers of neural networks by showing that the spatial features learned by depthwise separable convolutions remain generic \textbf{across all layers, domains, and architectures.} Our layer-wise analysis reveals that, in contrast to traditional CNNs, the transition from generic to class-specific filters is not observed in the deeper layers of depthwise separable CNNs. These findings suggest that the channel-wise information, rather than the spatial features, is responsible for task-specific adaptations in these networks. 
\end{abstract}

%Uncomment the following to link to your code, datasets, an extended version or similar.

\section{Introduction}

Understanding the mechanisms by which neural networks generalize across different tasks and datasets is a pivotal aspect of deep learning research~\cite{generalisationrethinkinh, neyshabur2017exploring}. Generalization, the ability of a model to perform well on unseen data, is often studied by evaluating a model's performance on new, unseen samples or its adaptability to novel domains. While many approaches focus on test accuracies and domain adaptation, in this work, we investigate the role of inner structural aspects of neural networks in generalization, particularly examining the properties of depthwise separable convolutional neural networks (DS-CNNs).

 %These convolutions manage to capture intricate spatial dependencies in the input data, which is crucial for high-level vision tasks. 
The first layer of traditional convolutional neural networks (CNNs) is well-documented to develop filters resembling Gabor functions or color blobs~\cite{alexnet}, indicative of their role in capturing basic edge and color information from the visual stimuli. However, as the network progresses into deeper layers, these patterns become more complex and less understood, given the increase in the number of channels and the entangled nature of spatial and channel representations in traditional CNNs.

The highly influential work of ~\cite{NIPS2014_375c7134} characterized the first layer filters of CNNs as "general," and extended its investigation to deeper layers, examining filter generality and specificity through innovative layerwise feature transfer experiments. They empirically demonstrated that when frozen filters from a dissimilar task were transferred, model performance degradation became progressively more severe as deeper layers were transferred. This led to the widely accepted conclusion that filters in deeper layers become increasingly specialized.

%On the other hand, they investigated the generality/specificity of filters in deeper layers through their famous layerwise feature transfer experiment. They empirically showed that frozen features transferred from a dissimilar task degrade performance, resulting in filters being specialized in deep layers. A conclusion that has been very influential and more than 10000 times cited. 

Depthwise separable convolutions are an efficient variant of the standard convolution operation, which decouples the learning of spatial features and channel-wise relationships~\cite{mobilenets, mobilenetv3}. This separation not only reduces computational complexity but also provides a unique lens through which the internal representation of spatial information can be inspected, even in deep layers of the network.

When probing the depthwise filters of trained models on ImageNet, one observes repeating patterns. Figure~\ref{fig:filtersamples} shows depthwise filters randomly sampled from the first, middle, and last layer of the trained ConvNeXt~\cite{convnext} Base and HorNet~\cite{hornet} Small models. The filters have common characteristics between the two different architectures. Recent study has shown that depthwise convolutions across different DS-CNN models trained on the ImageNet dataset are clusterable into distinct categories related to Gaussian functions and derivatives~\cite{babaiee2024unveiling, Babaiee_2024_WACV}.

 Inspired by these observations, our paper seeks to explore the possibility general filter sets being learned by depthwise separable convolutions across \textit{different domains, architectures, and model sizes}. 
 
 We hypothesize:

\begin{figure*}[t]
  \centering
  
  \begin{subfigure}{0.30\linewidth}
    \includegraphics[width=\linewidth, trim={0 5cm 0 0}, clip]{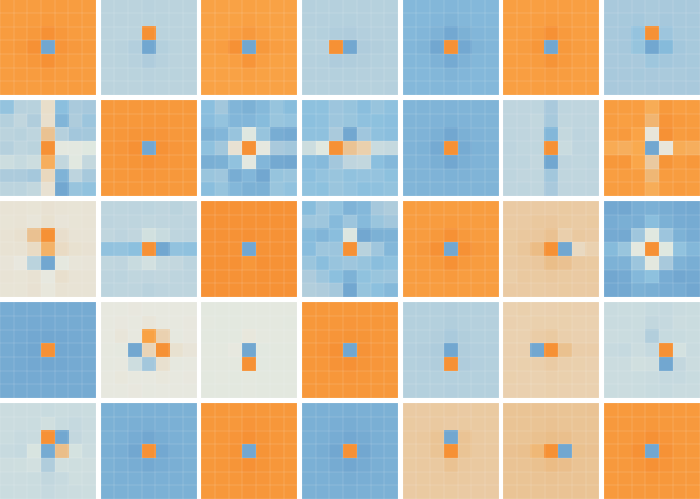}
    \caption{\textcolor{teal}{ConvNext} First Layer}
    \includegraphics[width=\linewidth, trim={0 5cm 0 0}, clip]{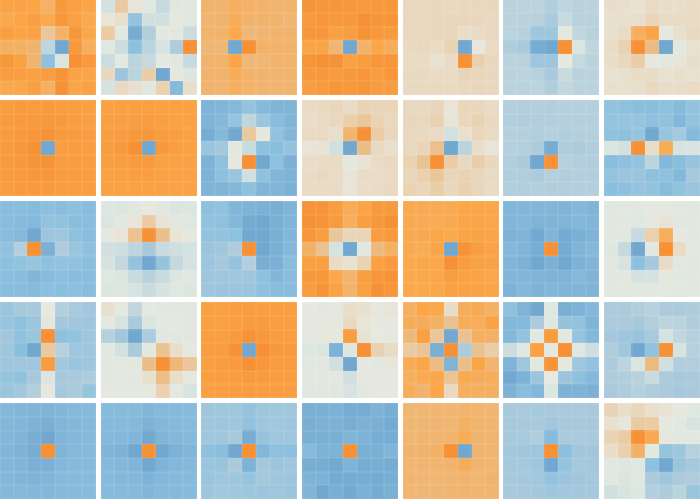}
    \caption{\textcolor{brown}{HorNet} First Layer}
  \end{subfigure}
  \hfill
  \begin{subfigure}{0.30\linewidth}
    \includegraphics[width=\linewidth, trim={0 5cm 0 0}, clip]{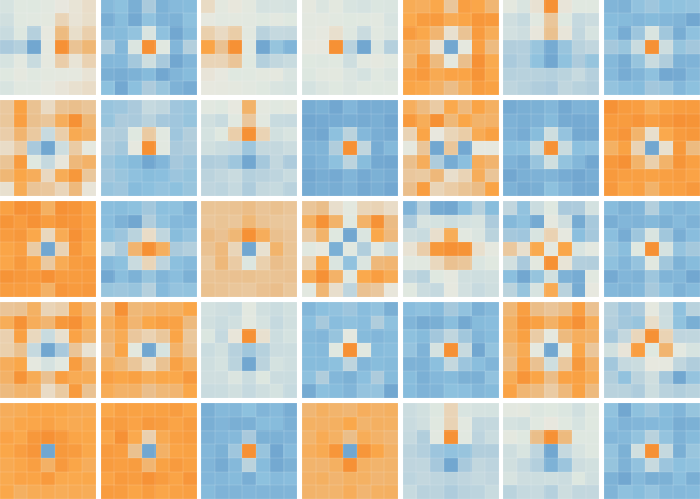}
    \caption{\textcolor{teal}{ConvNext} Middle Layer}
    \includegraphics[width=\linewidth, trim={0 5cm 0 0}, clip]{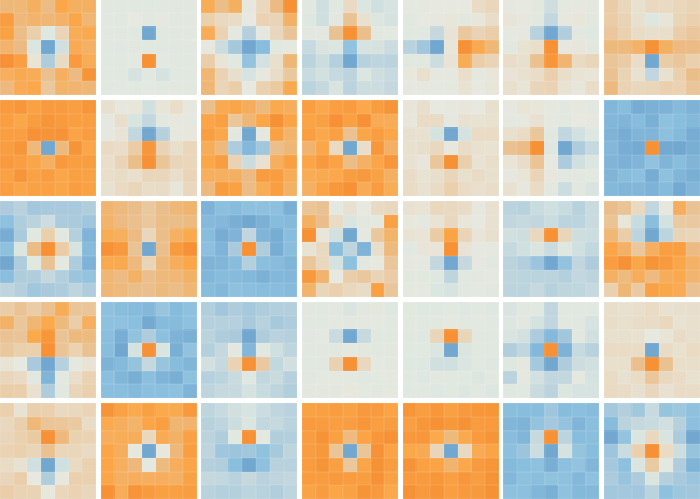}
    \caption{\textcolor{brown}{HorNet} Middle Layer}
  \end{subfigure}
  \hfill
  \begin{subfigure}{0.30\linewidth}
    \includegraphics[width=\linewidth, trim={0 5cm 0 0}, clip]{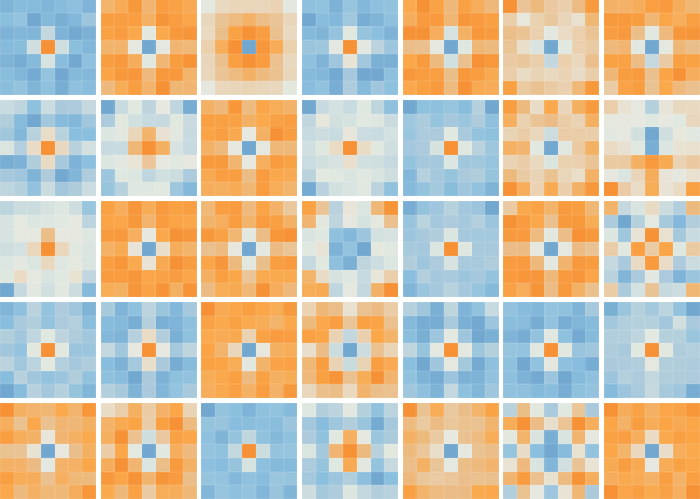}
    \caption{\textcolor{teal}{ConvNext} Last Layer}
    \includegraphics[width=\linewidth, trim={0 5cm 0 0}, clip]{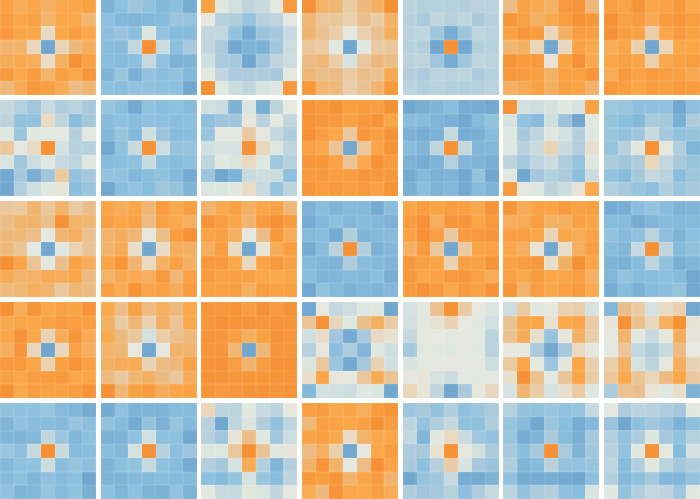}
    \caption{\textcolor{brown}{HorNet} Last Layer}
  \end{subfigure}
  \hfill
  
  \caption{Random depthwise filters sampled from the first, middle, and last layers of ConvNeXt Base and HorNet Small trained on ImageNet. Spatial features in DS-CNNs follow similar patterns regardless of the model architecture and layer.}
  \label{fig:filtersamples}
\end{figure*}

\textbf{The Master Key Filters Hypothesis.}
\textit{There exist master key filter sets that are general for visual data, and the depthwise filters in DS-CNNs tend to converge to these master key filters, regardless of the specific dataset, task, or architecture.}
 %\textit{The depthwise filters are general across datasets and architectures, and there is no transition to specialized filters even in deeper layers.}

To validate this hypothesis, we conduct a comprehensive series of experiments across ImageNet and various other datasets and domains. %It's crucial to recognize that merely demonstrating the clusterability of filters learned across different domains is insufficient to fully support our hypothesis. The variances of Gaussian filters or the proportions of clusters in each layer might still exhibit domain-specific arrangements, potentially indicating some level of specialization. To address this complexity and rigorously test our hypothesis, we have designed a set of targeted experiments:

\begin{enumerate}
    \item \textit{Semantically Divided ImageNet:} First, we repeat the well-known experiment from ~\cite{yosinski2015understanding}, by dividing ImageNet into "man-made" and "natural" classes that are semantically different from each other. We then transfer and freeze filters from the model trained on man-made classes to the model to be trained on natural classes. If our hypothesis is right, unlike ~\cite{yosinski2015understanding}, we should see no accuracy drops when transferring deeper layers.
    \item \textit{Cross Domain Transfer:} On a set of datasets from various domains, we transfer frozen filters from models trained on each dataset to the other, and investigate the performances.
    \item \textit{Cross Architecture Transfer:} We transfer the filters of a model architecture to another distinct model, trained on the same dataset.
    \item \textit{Cross Domain and Cross Architecture Transfer:} Finally, we transfer filters from a model with both a different architecture and also trained on a different dataset to another model to be trained on another dataset.
\end{enumerate}

\begin{figure*}[t]
  \centering
  
      \centering
    \includegraphics[width=\linewidth]{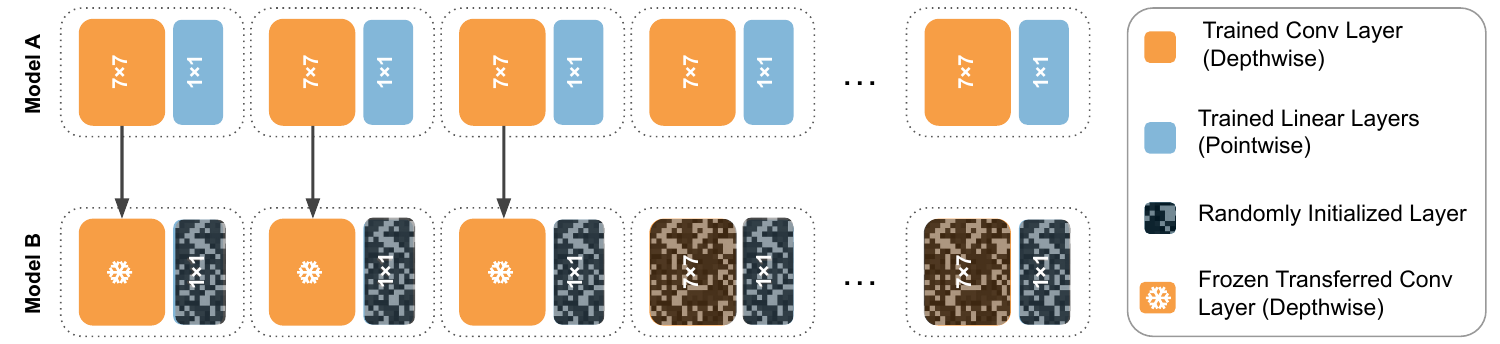}
    \caption{Overview of the experimental setup for depthwise filter transfers. Top: The base model-A
is trained on the source dataset-A.  Bottom: In the transfer model-B, the first n depthwise convolution layers of the network (in this example, n = 3)
are transferred and frozen from the base model-A, the rest of the layers are randomly
initialized, and then, they are trained on the target dataset-B. This
experiment tests the extent to which the filters on layer n are general or specific.}

  \label{fig:experimet-vis}
\end{figure*}

\section{Related Work}

\textbf{Generalization in Deep Learning.} Generalization has been a central theme in machine learning research for decades~\cite{neyshabur2017exploring}. The study of generalization seeks to understand how training methodologies, network architectures, and data diversity influence a model's ability to extend beyond its training regime~\cite{Goodfellow-et-al-2016}. Various theories, such as uniform convergence, margin theory, and algorithmic stability, have been proposed to explain generalization in machine learning. These frameworks often rely on different notions of model complexity, and corresponding generalization bounds quantify the relationship between the amount of data needed and the complexity measure. Despite significant theoretical advancements, the practical value and applicability of these theories remain a subject of ongoing debate in the research community~\cite{generalisationrethinkinh}.
Notably, Yosinski et al.~\cite{NIPS2014_375c7134} investigated the transferability of features in deep neural networks by transferring frozen filters from a CNN trained on half of ImageNet to a network to be trained on another half. They showed that transferring deeper than the third layer filters degrades performance, suggesting representation specificity in deep layers.

%Notably, Yosinski et al.'s seminal work "How transferable are features in deep neural networks?" highlights the nuances of feature transferability in CNNs. The authors empirically demonstrated that while transferring frozen filters from one part of the ImageNet dataset to a network trained on a different part can maintain performance across the first three layers, attempting to transfer deeper layers results in performance degradation. This finding suggests a high degree of representation specificity in deeper layers, indicating that early layers capture more generic features while deeper layers adapt to more dataset-specific features. In contrast to these findings, our work presents evidence that even deep layer depthwise filters can be effectively transferred across domains, suggesting a more universally applicable set of features than previously recognized.

\textbf{Depthwise Separable Convolutions.} Depthwise separable convolutions have gained popularity over traditional convolutions in recent years due to their computational efficiency and scalabilty ~\cite{mobilenets, mobilenetv3, efficientnet, mnasnet, MogaNet, convmixer, convnext}. They reduce parameter count and computational complexity by decoupling the spatial and channel computations. These layers have not only facilitated the development of lightweight, scalable models but have also been instrumental in probing the spatial feature extraction capabilities of CNNs.
A depthwise-separable convolution is an efficient alternative to standard convolutions in neural networks, splitting the operation into two simpler steps. First, a depthwise convolution applies a separate filter to each input channel independently, capturing spatial patterns within each channel. Mathematically, for an input $X$ with $C$ channels, this performs $C$ separate convolutions: $Y_c = X_c * K_c$,
where $K_c$ is the kernel for channel $c$. Second, a pointwise convolution ($1\times1$ convolution) combines information across channels by applying a $1 \times 1 \times C$ kernel to each spatial location, creating new feature maps: $Z = \sum_{c=1}^C Y_c W_c$,
where $W_c$ are the weights for each channel. This decomposition significantly reduces the number of parameters and computational cost compared to standard convolutions while maintaining similar expressiveness, making it particularly useful in mobile and edge computing applications.
Recent work demonstrates that depthwise convolutional kernels, across various DS-CNN models trained on the ImageNet dataset, exhibit recurring patterns that can be categorized into distinct groups~\cite{babaiee2024unveiling}.

\begin{table*}[h]
\renewcommand{\arraystretch}{1.2}
  \centering

  \begin{tabular}{@{}l|llll@{}}
     \textbf{Method} & Baseline & Transferred &  Shuffle Transferred & Only First 3 layers Transferred\\
    \midrule
    \textbf{Accuracy} & 86.9\% & 86.9\% & 86.2\% & 86.9\% \\
  \end{tabular}
    \caption{Accuracy Comparison of Different Filter Transfer Scenarios in ConvNeXt Tiny.}
\label{tab:shuffle}
\end{table*}
\textbf{Transfer Learning and Domain Adaptation.}
Transfer learning focuses on leveraging knowledge from one or more source tasks to improve learning in a related target task.
These approaches are particularly valuable in scenarios where labeled data for the target task is scarce or expensive to obtain. ~\cite{xu2024initializing} introduced a method for initializing smaller models by
transferring a subset of weights from a pre-trained larger model.
Transfer learning does not necessarily improve performance, as transferring knowledge from a dissimilar domain may not yield positive results~\cite{zhuang2020comprehensive}.  
In the realm of computer vision, many studies have investigated the factors that influence the transferability of models trained on ImageNet to other tasks. Kornblith et al.~\cite{kornblith2019better} found that while better ImageNet accuracy generally leads to better transfer performance, this relationship is not always consistent across different architectures and tasks. Additionally, the work by He et al.~\cite{he2019rethinking} challenges the conventional wisdom of ImageNet pre-training by demonstrating that training models from scratch on target datasets can achieve competitive or even superior performance compared to fine-tuning pre-trained models. 

In our work, we transfer \textit{the depthwise filters} and freeze them on a new domain in order to investigate the generality of the features learned across different datasets. Our findings suggest that the spatial features learned by depthwise filters possess a level of generality that allows them to be effectively transferred across different domains.
%Techniques range from fine-tuning pre-trained models (a common practice in deep learning) to complex domain adaptation strategies that seek to minimize the difference between source and target data distributions. 
%Transfer learning and domain adaptation techniques have been widely used to improve the generalization of deep learning models to new tasks and domains. Fine-tuning pre-trained models on new datasets has become a common practice, as it allows leveraging the learned features from large-scale datasets to improve performance on smaller, task-specific datasets [6]. Unsupervised domain adaptation techniques, such as adversarial learning [7] and self-training [8], have been proposed to bridge the gap between source and target domains without requiring labeled data in the target domain. These techniques aim to learn domain-invariant features that can generalize well to the target domain. In our work, we explore the transferability of learned depthwise filters across different domains, providing new insights into the nature of generalization in deep learning models.

\section{Generality of Spatial Features in DS-CNNs}

\begin{figure*}[t]
  \centering
  \begin{subfigure}{0.33\linewidth}
    \includegraphics[width=\linewidth]{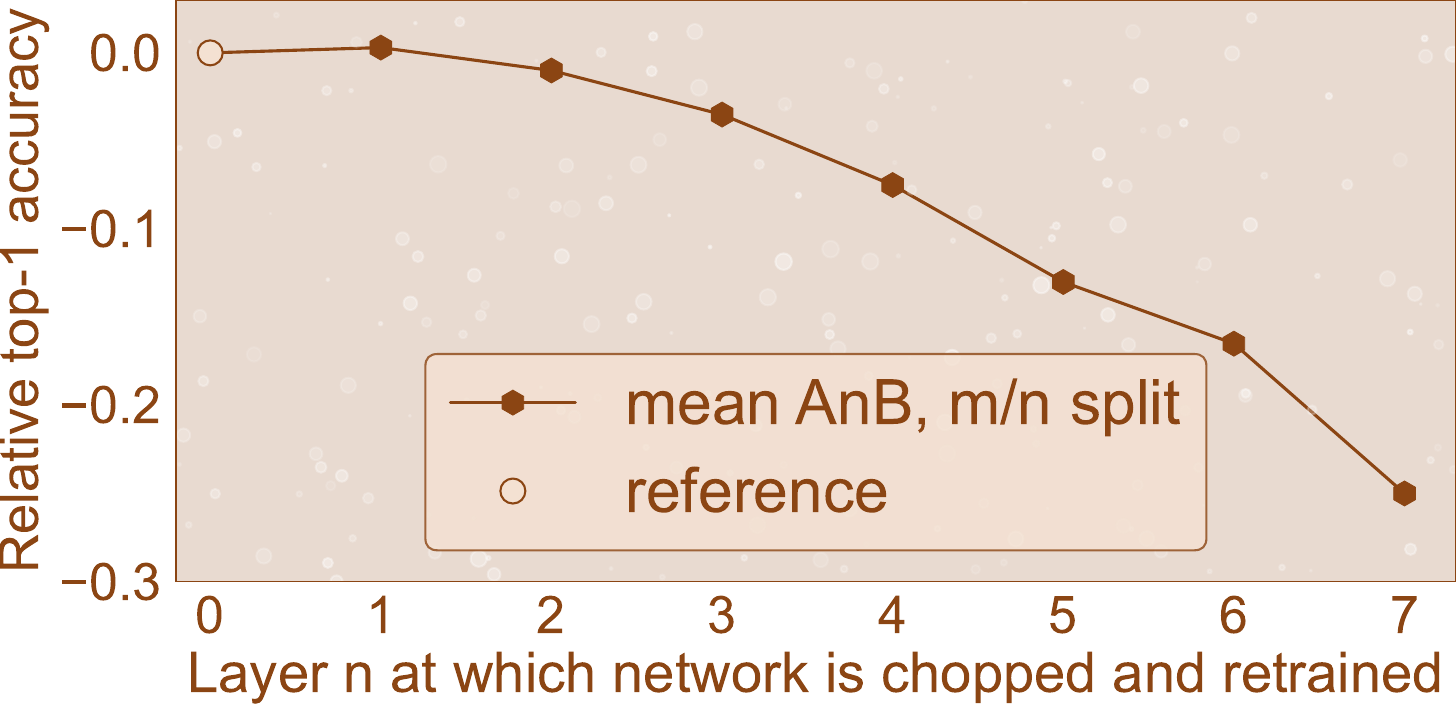}
    \caption{Plot Reproduced from~\cite{NIPS2014_375c7134} shows major performance degradation.}
    \label{subfig:2014}
  \end{subfigure}
  \hfill
  \begin{subfigure}{0.325\linewidth}
    \includegraphics[width=\linewidth]{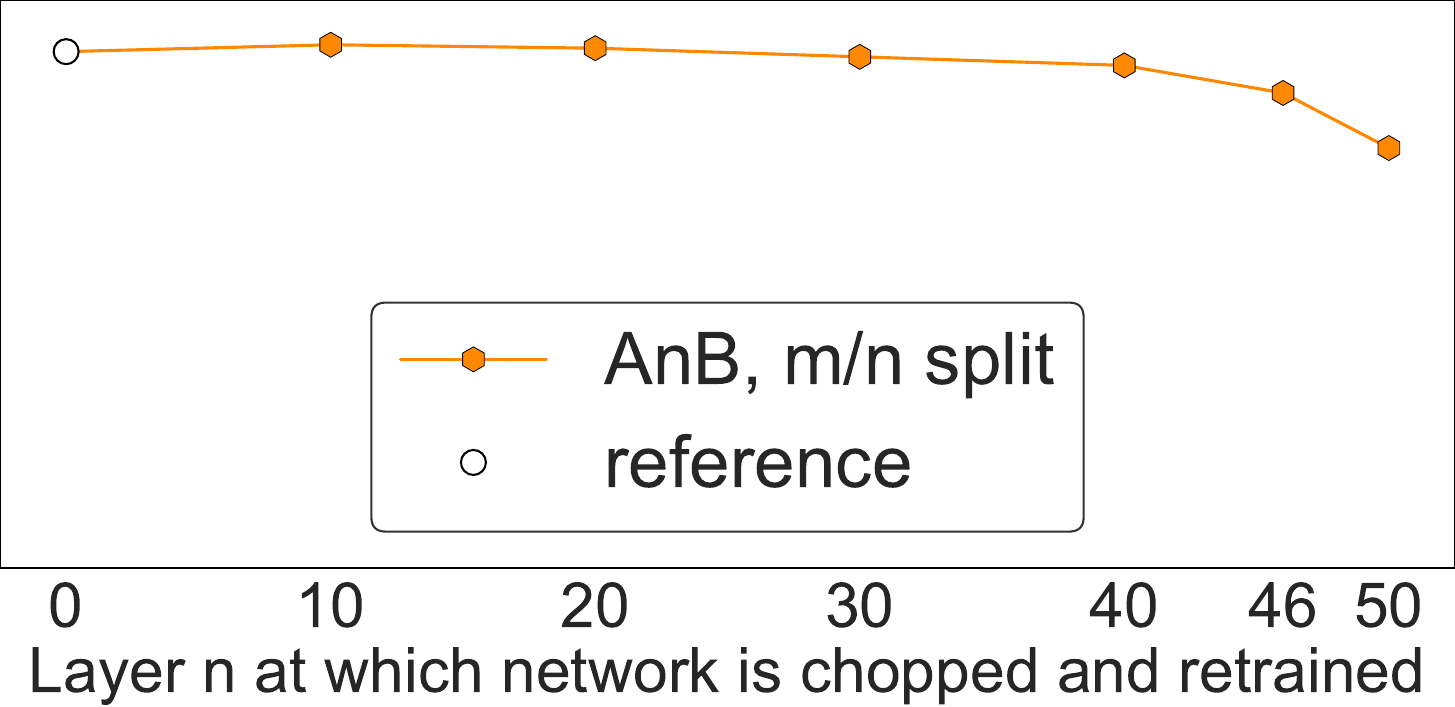}
    \caption{\textbf{ResNet50:} Model maintained 92.5\% of its original performance. }
    \label{subfig:resnet}
  \end{subfigure}
  \begin{subfigure}{0.325\linewidth}
    \includegraphics[width=\linewidth]{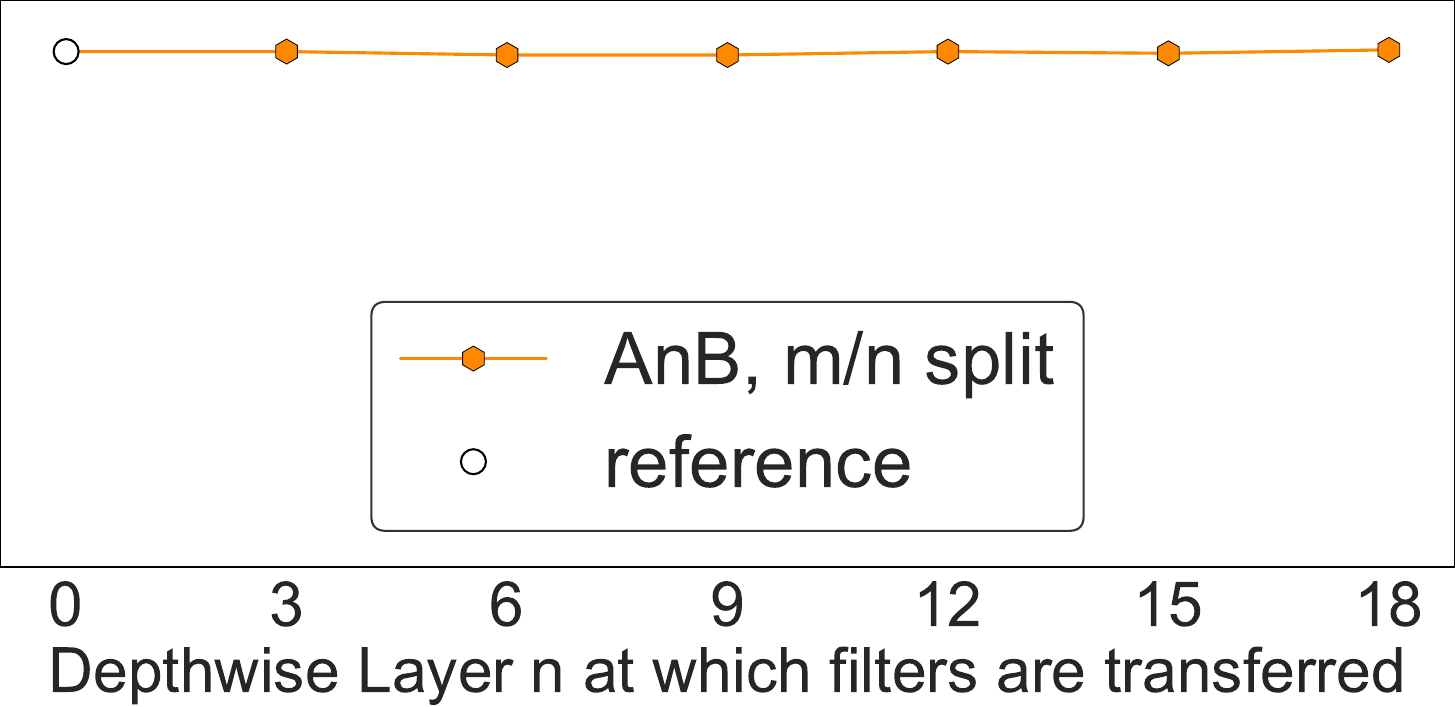}
    \caption{\textbf{ConvNeXt-T:} No performance drop, even at the last layer.}
    \label{subfig:convnext}
  \end{subfigure}
  \hfill
  
    \caption{This Figure replicates and extends the study by~\cite{NIPS2014_375c7134} using Resnets and DS-CNNs. ImageNet was split into man-made (m) and natural (n) classes. Networks A and B are trained on man-made and natural classes, respectively. The first $n$ layers are transferred from A to B, and this is denoted by AnB. The plots show relative accuracy to base models versus transfer depth. Each point indicates Network B's performance after transferring and freezing filters from A up to layer n, with the remaining layers trained on the natural subset.  Notably, depthwise filters exhibit high transferability across all layers, maintaining consistent performance regardless of transfer depth. This suggests a high degree of generality in depthwise convolutional filters, contrasting with 2014 experiment where performance degrades when transferring deeper layers between dissimilar domains. }

  \label{fig:manmade-natural}
\end{figure*}

%In this section, we investigate the generality of spatial features learned by DS-CNNs and their independence from the dataset and domain on which the network is trained. We draw inspiration from the work of Yosinski et al. [2], who define the degree of generality of a set of features learned on task A as the extent to which the features can be used for another task B. However, it is crucial to note that this definition depends on the similarity between tasks A and B.

%Driven by the intriguing phenomenon of clusterable filter patterns in depthwise convolutions within DS-CNNs trained on the ImageNet dataset~\cite{babaiee2024unveiling}, 

Our primary inquiry centers on whether the spatial features learned by DS-CNNs are universally applicable across various datasets and domains. Drawing on the conceptual framework presented in~\cite{NIPS2014_375c7134}, the generality of learned features can be defined by their utility when applied to tasks beyond their original training purpose. Specifically, examining how effectively these features perform when transferred from their initial training task to a different target task. The feasibility of such transfer relies significantly on the similarity between the source and target tasks.

To rigorously test our hypothesis, we engage in an extensive experimental process where we transfer and freeze the depthwise filters from models trained on different datasets.

\subsection{Revisiting Semantically Divided ImageNet}
%In this section, we replicate the experiment by ~\cite{NIPS2014_375c7134} for the depthwise convolutional filters in DS-CNNs.

In this section, we replicate the experiment by \cite{NIPS2014_375c7134} on convolutional filter transferability across ImageNet subsets (man-made vs. natural objects) on DS-CNNs. This division creates maximally dissimilar subsets within the ImageNet dataset.

As demonstrated in Figure~\ref{fig:experimet-vis}, we transferred the depthwise filters from the first n layers of the model trained on the man-made subset to a new ConvNeXt tiny model. These transferred layers were then frozen, and the model was trained on the natural subset. Figure~\ref{subfig:convnext} illustrates the performance results, and Figure~\ref{subfig:2014} shows the exact results re-plotted from~\cite{NIPS2014_375c7134}. Contrary to~\cite{NIPS2014_375c7134}, on ConvNeXt, transferred filters perform comparably to those trained directly on the natural subset, with no substantial performance trend as the number of transferred layers increases.

To evaluate the breadth of filter generality, we conducted experiments comparing three distinct transfer scenarios against our baseline accuracy. These scenarios included: (1) standard transfer of all filters (Figure~\ref{subfig:convnext}), (2) random shuffling of filters across layers, and (3) a restricted transfer where only the first three layers' filters were used and then repeated throughout the remaining layers. Table~\ref{tab:shuffle} shows the results. Surprisingly, even extreme scenarios like retaining only the first 3 layers showed no significant accuracy drop. These results strongly support the high generality of depthwise convolution filters across layers.

These findings raise an important question: Is the enhanced transferability of deeper filters in DS-CNNs, compared to the traditional CNN studied by Yosinski et al., due to the ConvNeXt model's depthwise separable architecture, or do modern training techniques play a crucial role in helping filters learn more generalizable patterns? To investigate this, we conducted the same experiment using ResNet50, a traditional CNN architecture trained with modern methods. The results, shown in Figure 3, reveal that the model stays robust, maintaining 92.5\% of it's performance, even when all of it's 49 convolutional layers are transferred.

\begin{figure}[]
    \includegraphics[width=0.695\linewidth]{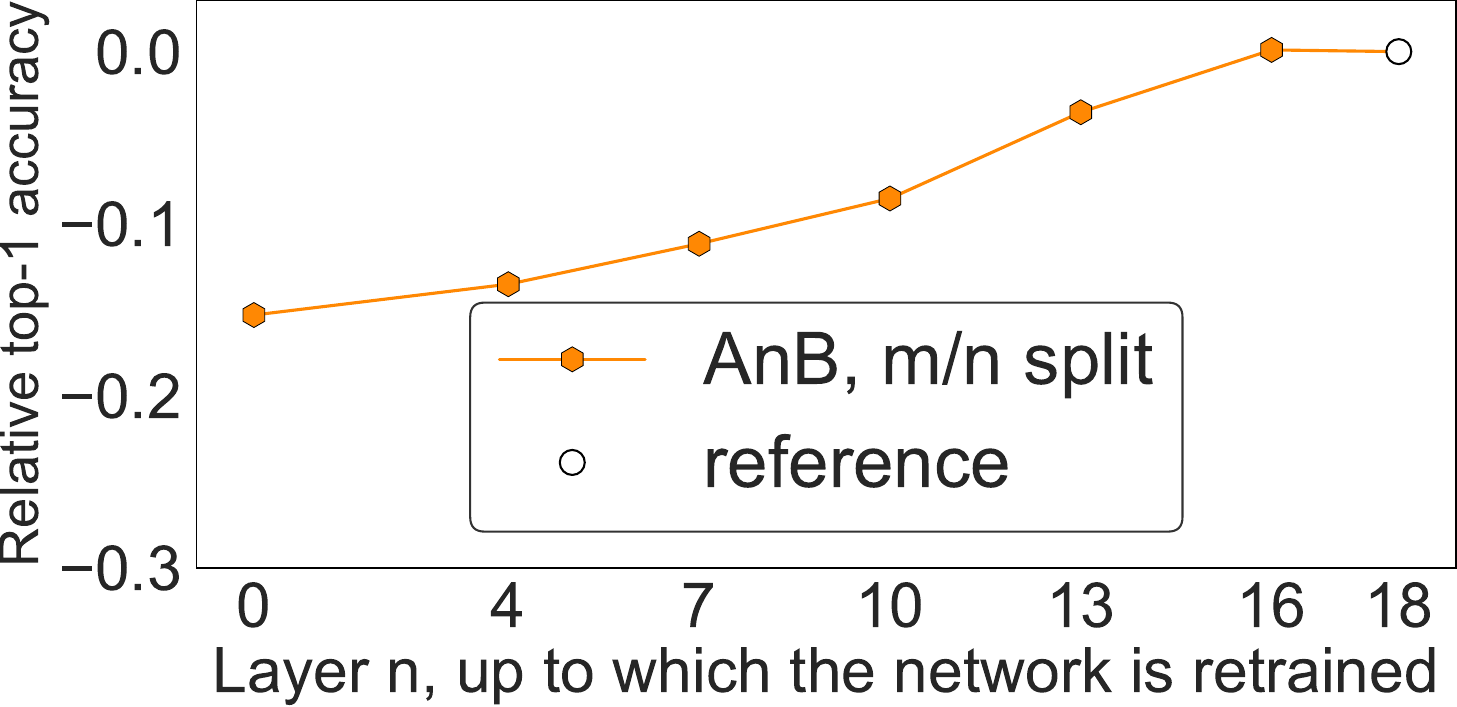}
    \caption{\textbf{ResNet18}: Figure~\ref{fig:manmade-natural} experiment, but reversed. Freezing layers from the \textbf{\emph{end} to \emph{n}} while training \emph{earlier} layers. Following same reasoning as Yosinski's, these results would suggest later layers are general while initial layers are special.}
\label{fig:reverse}
\end{figure}

\begin{figure}
    \centering
    \includegraphics[width=\linewidth]{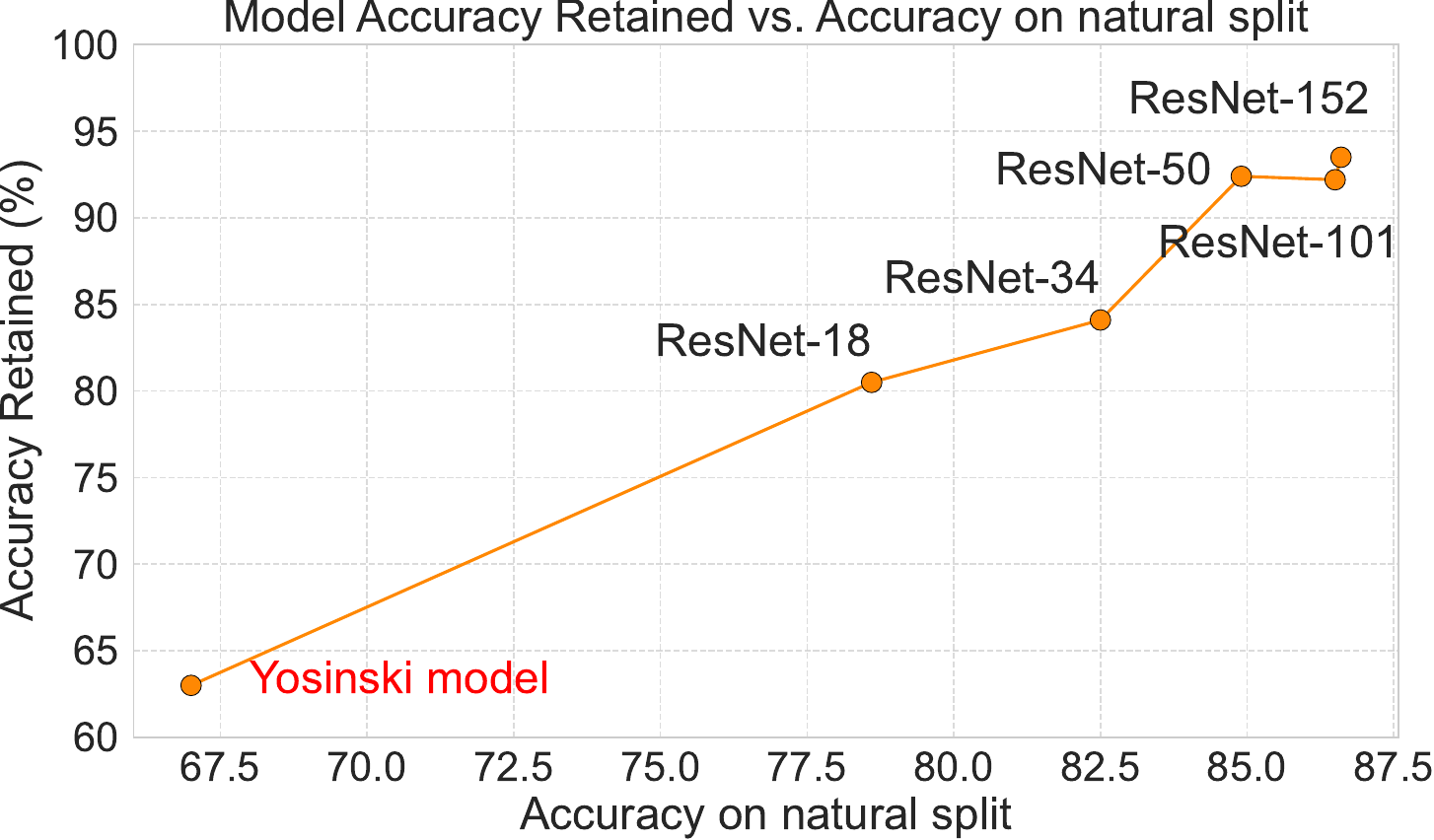}
    \caption{Accuracy retention comparison across ResNet architectures. Deeper networks (ResNet-50/101/152) maintain substantially higher accuracy after transfer compared to shallower networks (ResNet-18/34) and Yosinski's model.}
\label{fig:resnet_comparison}
\end{figure}

The contrasting results between Figure~\ref{subfig:2014} and Figure~\ref{subfig:resnet} raise another question: Could the performance retention observed in ResNet model be attributed to the residual connections, which were absent in Yosinski's model? Our experimental evidence suggests otherwise. Figure~\ref{fig:resnet_comparison} illustrates the accuracy retention across different ResNet architectures following complete layer transfer. We define accuracy retention as the percentage of original accuracy maintained after transfer (e.g., if accuracy decreases from 80\% to 70\%, the retained accuracy is 70/80=87.5\%).

The data reveals a notable pattern: smaller ResNet models exhibit significantly lower accuracy retention, approaching levels similar to those observed in Yosinski's model. This finding has two important insights. First, it challenges the idea that residual connections are strongly responsible for better transfer performance, as smaller ResNets having these connections show poor retention. Second, it also cast new doubts about the "specialized later layers" notion. If layer specialization were an inherent property of CNNs, we would expect to observe increased specialization in models with more layers and, crucially, more pronounced specialization in higher-accuracy models. However, our results demonstrate the opposite trend: deeper ResNet architectures with higher accuracy exhibit more generalized behavior in their later layers.

To further challenge the notion of layer specialization, we conducted the reverse of Yosinski's experiment—freezing layers from the end to n while training earlier layers in  Figure~\ref{fig:reverse}. Following the same interpretation as Yosinski's framework, these results would paradoxically suggest that later layers are general while initial layers are specialized.

\subsection{Cross Domain Transfer}

This section examines the cross-domain transferability of depthwise separable convolutional filters using a diverse set of datasets varying in size and domain. We aim to assess the generalizability of these filters across disparate datasets.

\begin{table}[h]

  \setlength{\tabcolsep}{7pt} % Reduce space between columns
  \renewcommand{\arraystretch}{1.0}
  \centering
  \begin{tabular}{@{}l|lll@{}}
    \textbf{Dataset} & \textbf{Classes} & \textbf{Train} & \textbf{Test} \\
    \midrule
    ImageNet & 1000 & 1.2 m & 50 k \\
    Food 101 & 101 & 75 k & 25 k \\
    Sketch & 345 & 50 k & 20 K \\
    CIFAR-10 & 10 & 50 k & 10 k  \\
    STL-10 & 10 & 5 k & 8 k \\
    Oxford-IIIT Pets & 37 & 4 k & 3 k  \\
    Oxford 102 Flowers & 102 & 2 k & 6 k \\
  \end{tabular}
  \caption{Dataset information and sample sizes, in training set size descending order.}
    \label{tab:datasets}
\end{table}

\begin{table*}[t]

  \setlength{\tabcolsep}{9pt} % Reduce space between columns
  \centering
  \begin{tabular}{@{}l|lllllll@{}}
    \textbf{Dataset}     & 
    ImageNet & Food & Sketch & Cifar10 & STL10 & Pets & Flowers  \\ 
    \midrule
    % \textbf{ImageNet} & 76.1\% & &  & & & \\ 
    \textbf{Accuracy} & 76.1\% & 87.6\% & 66.6\% & 96.9\%  & 80.4\% & 36.3\%& 66.0\%\\ 
  \end{tabular}
  \caption{ConvNeXt Femto model accuracy on the datasets in our benchmark.}
    \label{tab:femto_accs}
\end{table*}

\begin{table*}[t]

  \setlength{\tabcolsep}{9pt} % Reduce space between columns
    \renewcommand{\arraystretch}{1.5} % Adjust the row height (1.5 times the default height)

  \centering
\begin{tabular}{@{}l|lllllll@{}} \diagbox{\textbf{Target}}{\textbf{Source}} & ImageNet & Food & Sketch & Cifar10 & STL10 & Pets & Flowers \\ \midrule Food & \textcolor{darkgreen}{+}0.3\% & \cellcolor{darkred!10}\textcolor{darkred!10}{+}87.3\% & \textcolor{darkred}{-}1.2\% & \textcolor{darkred}{-}3.7\% & \textcolor{darkred}{-}5.5\% & \textcolor{darkred}{-}9.1\%& \textcolor{darkred}{-}8.2\%\\ Sketch & \textcolor{darkgreen}{+}0.5\% & \textcolor{white}{-}0.0\% & \cellcolor{gray!10}\textcolor{gray!10}{+}66.6\% & \textcolor{darkred}{-}1.4\% & \textcolor{darkred}{-}2.5\% & \textcolor{darkred}{-}5.0\% & \textcolor{darkred}{-}5.6\%\\ Cifar10 & 0.0\% & -0.1\% & +0.1\% & \cellcolor{darkgreen!10}\textcolor{darkgreen!10}{+}97.1\% & \textcolor{darkred}{-}0.7\% & \textcolor{darkred}{-}1.1\% & \textcolor{darkred}{-}1.3\% \\ STL10 & \textcolor{darkgreen}{+}0.5\% & \textcolor{darkgreen}{+}1.2\% & \textcolor{darkgreen}{+}1.9\% & \textcolor{darkgreen}{+}2.5\% & \cellcolor{darkgreen!10}\textcolor{darkgreen!10}{+}82.7\% & \textcolor{darkred}{-}3.4\% & \textcolor{darkred}{-}3.4\% \\ Pets & \textcolor{darkgreen}{+}3.6\% & \textcolor{darkgreen}{+}9.5\% & \textcolor{darkgreen}{+}11.5\% & \textcolor{darkgreen}{+}4.4\% & \textcolor{darkgreen}{+}7.6\% & \cellcolor{darkgreen!10}\textcolor{darkgreen!10}{+}52.4\% & \textcolor{darkred}{-}7.2\% \\ Flowers & \textcolor{darkgreen}{+}4.1\% & \textcolor{darkgreen}{+}2.5\% & \textcolor{darkgreen}{+}2.3\% &\textcolor{darkgreen}{+}4.2\% & \textcolor{darkgreen}{+}2.7\% & -0.1\% & \cellcolor{darkgreen!10}\textcolor{darkgreen!10}{+}69.1\% \\ \end{tabular}
  \caption{Accuracy of the ConvNeXt Femto model on the target dataset, with frozen \textbf{depthwise filters} transferred from the models trained on source datasets. The diagonal shows the results of the models with "selffer" frozen depthwise filters. The datasets are ordered based on descending training set size. The cell colors red, green, and gray show a decrease, increase, or no change in selffer accuracy compared to the original accuracy, respectively. Arrows indicate relative accuracy ($\geq 0.1$) compared to the selffer models in each row.}
    \label{tab:results_main}
    
\end{table*}
\textbf{Datasets.} We evaluate the transferability of depthwise filters across six diverse datasets: Food 101~\cite{food101}(food images), Sketch~\cite{domainnet} (hand-drawn object sketches from DomainNet), CIFAR-10~\cite{cifar10} (generic images of vehicles and animals), Oxford Flowers~\cite{flowers} (various flower species), Oxford Pets~\cite{pets} (cat and dog breeds), and STL-10\cite{stl10} (mix of animals and objects). 
 The diverse datasets present distinct visual features to test filter transferability. Details of these datasets are summarized in Table~\ref{tab:datasets}.

\textbf{Base Model.} Our experiments utilize the ConvNeXt Femto model\cite{convnext}. We train the model for 300 epochs in each run, maintaining uniform hyper-parameters across all training instances.

\textbf{Experimental Setup.} Our experimental process is as follows:

\begin{enumerate}
    \item Train the ConvNeXt Femto model on each of the six datasets separately.
    \item For each pair of datasets, transfer all depthwise filters across all layers from the source domain to the network that will be trained on the target dataset and freeze them.
    \item Train the model on the target dataset with the transferred and frozen filters.
    \item Additionally, for each dataset, transfer the filters of the model trained on ImageNet using the same process.
\end{enumerate}

The models with transferred filters have the advantage of already trained filters. Hence, for a fair comparison of the transferred filters from other domains with the original dataset's accuracy, for each dataset, we also transfer the filters from the model once already trained on it, freeze them, and train the model for another 300 epochs - a baseline termed "selffer" (self-transferred). This process results in 42 different training configurations. The results of these experiments are presented in Tables \ref{tab:femto_accs} and \ref{tab:results_main}. 

%The models with transferred filters have the advantage of already trained filters. Hence, for a fair comparison of the transferred filters from other domains with the original dataset's accuracy, for each dataset, we also transfer the filters from the model once already trained on it, freeze them, and train the model for another 300 epochs. 

\textbf{Results.} The formatting of Table~\ref{tab:results_main} is particularly illustrative. The table is arranged such that datasets are ordered by their size, with ImageNet, featuring over a million samples, positioned at the forefront, and the Oxford Flowers dataset, which has approximately 2000 samples, at the other end. The diagonal cells denote the "selffer" accuracies. Performance metrics are color-coded to enhance readability and interpretability. %

%This process results in 30 different training configurations. The results of these experiments are presented in Tables \ref{tab:femto_accs} and \ref{tab:results_main}. The diagonal of Table~\ref{tab:results_main} marked in gray shows the own-transfer accuracies. The datasets in rows and columns are sorted based on their size, ImageNet with more than 1 million training samples is the largest and comes first and the flowers with around 2000 has the least number of training samples. We take the own-transfer performance as the base, and the accuracies of filter transfers lower than the own transfer are marked in red, and the ones resulting in higher accuracies are in green. The cases with smaller than 0.1\% change in accuracy remain white.

These results reveal key insights, which we will discuss further.

\textbf{Asymmetric Transfer Effects and Dataset Size:} The pattern revealed in Table~\ref{tab:results_main} challenges conventional assumptions about domain-specificity in filter transfer. The predominantly red arrows in the upper triangle and green arrows in the lower triangle, when datasets are sorted by size, indicate that filters from models trained on larger datasets consistently outperform those from smaller ones. This improvement persists regardless of domain similarity between source and target datasets, suggesting that increased data variety leads to more universally applicable filters. Moreover, we never observe mutual negative impact when transferring filters between datasets—a finding that contradicts what might be expected if filters were highly domain-specific. This asymmetric pattern suggests that depthwise filters develop general capabilities that become more robust with increased training variety rather than becoming narrowly specialized to specific domains.

%\textbf{Effect of Data Variety on Transfer Effectiveness:} In Table~\ref{tab:results_main}, the upper triangle is predominantly with red arrows, while the lower triangle is generally with green arrows. Given that the datasets in this table are sorted based on size, this observation provides us with valuable insights: the filters of models trained on larger datasets tend to outperform those trained on smaller ones. Specifically, transferring filters from models trained on datasets with higher sample varieties almost always improves the performance. This improvement is consistent regardless of the domain similarity between the source and target datasets, suggesting that larger datasets may help models develop more universally applicable filters. Larger datasets provide a more comprehensive set of examples, which allows the model to better learn and generalize the underlying patterns and features. As a result, models trained on larger datasets can converge more effectively to general filters, enhancing the overall transfer effectiveness and robustness of the filters when applied to different domains.

\begin{table*}[t]

  \setlength{\tabcolsep}{9pt} % Reduce space between columns
    \renewcommand{\arraystretch}{1.5} % Adjust the row height (1.5 times the default height)

  \centering
\begin{tabular}{@{}l|lllllll@{}} \diagbox{\textbf{Target}}{\textbf{Source}} & ImageNet & Food & Sketch & Cifar10 & STL10 & Pets & Flowers \\ \midrule Food & $\textcolor{darkred}{-}$0.9\% & \cellcolor{darkred!10}$\textcolor{darkred!10}{-}$63.6\% & $\textcolor{darkred}{-}$1.1\% & $\textcolor{darkred}{-}$0.9\% & $\textcolor{darkred}{-}$2.9\% & $\textcolor{darkred}{-}$3.9\% & $\textcolor{darkred}{-}$4.1\%\\ Sketch & $\textcolor{darkred}{-}$2.4\% & $\textcolor{darkred}{-}$2.7\% & \cellcolor{darkred!10}$\textcolor{darkred!10}{-}$47.7\% & $\textcolor{darkred}{-}$2.8\% & $\textcolor{darkred}{-}$4.2\% & $\textcolor{darkred}{-}$4.4\% & $\textcolor{darkred}{-}$4.6\%\\ Cifar10 & $\textcolor{darkred}{-}$0.7\% & $\textcolor{darkred}{-}$0.9\% & $\textcolor{darkred}{-}$0.2\% &\cellcolor{darkred!10}$\textcolor{darkred!10}{-}$87.7\% & $\textcolor{darkred}{-}$0.3\% & $\textcolor{darkred}{-}$0.2\% & $\textcolor{darkred}{-}$1.7\% \\ STL10 & $\textcolor{darkred}{-}$0.4\% & $\textcolor{darkred}{-}$3.2\% & $\textcolor{darkred}{-}$3.2\% & $\textcolor{darkred}{-}$0.1\% & \cellcolor{darkred!10}$\textcolor{darkred!10}{-}$70.5\% & $\textcolor{darkred}{-}$0.5\% & $\textcolor{darkred}{-}$3.0\% \\ Pets & $\textcolor{darkred}{-}$11.8\% & $\textcolor{darkred}{-}$2.4\% & $\textcolor{darkred}{-}$7.0\% & $\textcolor{darkred}{-}$7.3\% & $\textcolor{darkred}{-}$4.5\% & \cellcolor{darkred!10}$\textcolor{darkred!10}{-}$43.6\% & $\textcolor{darkred}{-}$10.4\% \\ Flowers & $\textcolor{darkred}{-}$4.1\% & $\textcolor{darkred}{-}$3.0\% & $\textcolor{darkred}{-}$3.0\% &$\textcolor{darkred}{-}$4.6\% & $\textcolor{darkred}{-}$1.0\% & $\textcolor{darkred}{-}$1.0\% & \cellcolor{darkred!10}$\textcolor{darkred!10}{-}$57.0\% \\ \end{tabular}
  \caption{Accuracy of the ConvNeXt Femto model on the target dataset, with with frozen \textbf{pointwise filters} transferred from the models trained on source datasets. The diagonal shows the results of the models with "selffer" frozen pointwise filters. The datasets are ordered based on descending training set size. The cell colors red show a decrease in selffer accuracy compared to the original accuracy, respectively. Arrows indicate relative accuracy compared to the original models in each row.}
    \label{tab:results_pw}

\end{table*}

%\textbf{Asymmetry in Transfer Impact:} Another compelling observation is the absence of mutual negative impact in performance when filters are transferred between two datasets. This is notable because, under the premise of domain-specificity, one might expect that filters optimized for a particular domain (such as foods or pets) would not only fail to improve but actively degrade performance when applied to a distant domain. Yet, our results do not support this; we do not record instances where filter transfer between two datasets results in decreased performance in both directions. This challenges the notion of high domain-specificity in deeper layers and supports the idea of a more generalized capability within the depthwise filters.

\subsubsection{Is There a Transition from Generic to Class-specific Filters in Deeper Layers?}

In the experiments in Table~\ref{tab:results_main}, we transferred all the filters from all layers to the new models on new domains. To investigate the layer trends, we perform a similar study to the previous section.

We continue to use ConvNeXt Femto as our base model. For the source and target datasets, we use Food 101 and Oxford Pets.
Starting from a ConvNeXt Femto model with all layers trained on the Food 101 dataset, we iteratively transfer layers, similar to the procedure shown in Figure~\ref{fig:experimet-vis}.

We also repeat the same process by selfferring filters from a ConvNeXt Femto model trained on the Oxford Pets dataset.

\begin{figure}
    \centering
    \includegraphics[width=0.8\linewidth]{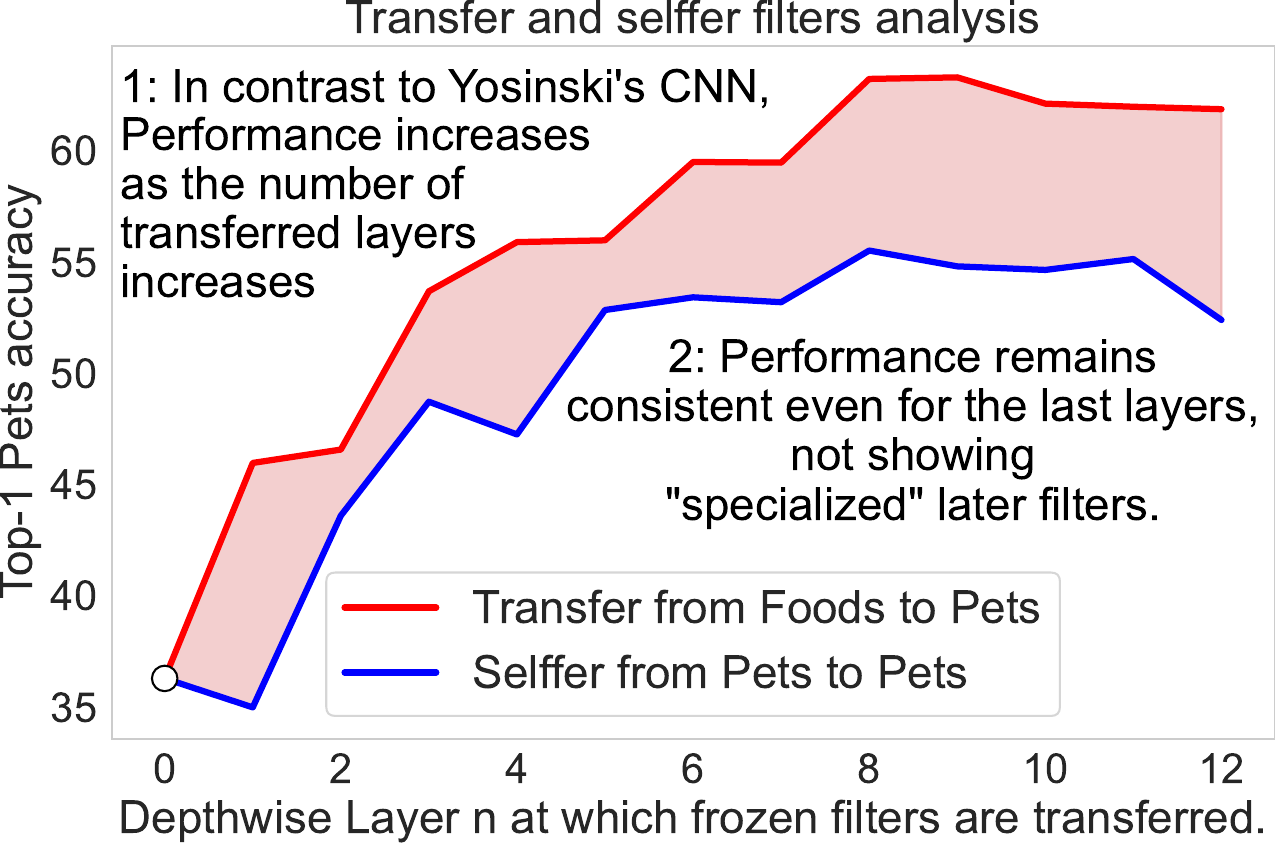}
    \caption{Transferring filters from all layers of a model trained on Foods, improves the performance of the model trained on the distant dataset Pets. Depthwise filters learn general features even in the latest layer, with stark contrast to the task-specialized filters phenomena in Yosinski's CNN.}
    \label{fig:acc_curve}
\end{figure}

 Figure~\ref{fig:acc_curve} shows the accuracy change for both the Food 101-to-Oxford Pets and Oxford Pets-to-Oxford Pets transfer of filters, with respect to the layer number where the source model is chopped. Remarkably, the results demonstrate that the more layers we transfer the filters from, the better the performance becomes. After transferring filters from 8 layers, the performance does not change considerably. These findings too, stand in stark contrast to those previously observed in~\cite{NIPS2014_375c7134}, where the performance degraded as more layers of the model trained on the far domain were transferred, especially after the first three layers.

\subsubsection{What About Pointwise Convolutions, Are They Specialized?}

The results thus far indicate that depthwise filters exhibit significant generality. This raises an intriguing question: If DS-CNNs extract features hierarchically and transition to specialized features, are the pointwise convolutions responsible for this specialization? To address this, we conducted another cross-domain experiment, transferring only the pointwise layers while training the remaining model weights. Table~\ref{tab:results_pw} presents the results of these experiments for each pair of datasets.

Surprisingly, transferring pointwise filters consistently decreased accuracy compared to the original model, even in selffer experiments. While improved or maintained accuracy during transfers can suggest filter generality, \textit{the accuracy decreases don't necessarily prove pointwise filter specialization.} This is particularly evident given that \textit{pointwise filters} transferred from the same dataset also showed significant drops, in contrast to selfferred \textit{depthwise filters}, which generally maintained or improved performance.

The performance degradation observed in these experiments may be attributed to optimization challenges related to splitting networks between co-adapted neurons. This phenomenon, termed "fragile co-adaptation" by ~\cite{NIPS2014_375c7134}, suggests that freezing transferred layers may create a loss landscape that hinders optimal filter learning. This difficulty is underscored by the fact that selfferred pointwise filters suffer similarly to those transferred from other domains. Upon examining the filters learned in these experiments, we observed notably noisier patterns, further indicating potential convergence issues.

\subsection{Cross-Architecture Transfer}

To further investigate the transferability of depthwise filters, we extend our experiments to include different model sizes and architectures, while using the ImageNet dataset as the source domain.

\textbf{Experimental Setup.} For these experiments, we maintain our base model as ConvNeXt Femto and use the Oxford Pets dataset as the target domain. As source models, we use different sizes from the ConvNeXt family (femto, tiny, and large) and introduce another architecture family, HorNet~\cite{hornet}. HorNet has substantially different blocks compared to ConvNeXt, with recursive gated convolutions. By including HorNet, we can evaluate the transferability of filters from a diverse set of model sizes and architectures.

When a source model is larger than the ConvNeXt Femto or has a different number of channels in its layers, we stack all the depthwise filters in the model and transfer them from the beginning of the stack to the ConvNeXt Femto. These transferred filters are then frozen and used to train the model on the Oxford Pets dataset.

\textbf{Results.} The results of these experiments are presented in Table~\ref{tab:pets_transfers}. Transferring filters from larger variants of ConvNeXt trained on ImageNet leads to better accuracy improvements compared to the smaller variant. Interestingly, the transferred filters from HorNet perform exceptionally well, despite being from a completely different model architecture and size.

\begin{table}[h]
\renewcommand{\arraystretch}{1.2}

\setlength{\tabcolsep}{3.5pt} % Reduce space between columns
\centering
\begin{tabular}{@{}l|ccc|ccc@{}}
& \multicolumn{3}{c|}{ConvNeXt} & \multicolumn{3}{c}{Hornet} \\
\textbf{Source Model} & Femto & Tiny & Large & Tiny & Small & Large \\
\midrule
\textbf{Accuracy} (\%)& \cellcolor{darkgreen!10}56.0 & \textcolor{darkgreen}{+}11.0 & \textcolor{darkgreen}{+}10.5 & \textcolor{darkgreen}{+}4.3 & \textcolor{darkgreen}{+}7.3 & \textcolor{darkgreen}{+}2.8\\
\end{tabular}
\caption{Accuracy of ConvNeXt Femto on the Oxford Pets dataset with transferred filters from different model architectures and sizes trained on ImageNet.}
\label{tab:pets_transfers}

\end{table}

%These results demonstrate the broad applicability and transferability of depthwise filters in DS-CNNs. Filters from larger models and diverse architectures consistently improve performance when transferred, indicating that learned spatial features are independent of dataset, domain, model architecture, and size. DS-CNNs' ability to capture fundamental spatial patterns explains the generality of these learned filters, making them valuable for improving performance in scenarios with limited data or computational resources.

These results demonstrate that depthwise filters learned by DS-CNNs are highly transferable, with successful transfer even from various architectures despite structural differences. The consistent performance improvements suggest that these spatial features generalize well across datasets, domains, and model architectures.

\subsection{Cross-Domain and Cross-Architecture Transfer}

To further demonstrate the generality of the depthwise filters, we extend our experiments by considering both different domains and different architectures simultaneously. While the ImageNet dataset is large and may already contain features useful for classifying pets, we aim to investigate the transferability of filters from a more distant domain. For this purpose, we choose the Food 101 dataset as the source domain, which consists of closeup photos of food on plates or table settings. In contrast, the Oxford Pets dataset, which serves as the target domain, contains images of cat and dog breeds in various settings, such as indoors or outdoors on grass. By selecting these two datasets, we can assess the effectiveness of filter transfer between domains that have minimal common features.

\textbf{Experimental Setup.} We first train the HorNet Tiny model on the Food 101 dataset. We then transfer the depthwise convolutional filters from the trained HorNet Tiny model to the ConvNeXt Femto model, which is subsequently trained on the Oxford Pets dataset with frozen filters. In this scenario, both the dataset domain and the model architecture of the source and target models are different, providing a rigorous test for the generality and transferability of the learned filters.

\textbf{Results.} The results of this experiment are presented in Table \ref{tab:cross_domain_arch}. Remarkably, the ConvNeXt Femto model trained on the Oxford Pets dataset with transferred filters from the HorNet Tiny model trained on the Food 101 dataset achieves an accuracy of 55.5\%, with a 3.1\% increase compared to the selfferred baseline. This result is particularly impressive considering the significant differences between the source and target domains, as well as the distinct model architectures.

\begin{table}[t]
\renewcommand{\arraystretch}{1.2}
  \centering

  \begin{tabular}{@{}l|ll@{}}
    Source Model & ConvNext-F on pets & HorNet-T on Foods \\
    \midrule
    \textbf{Accuracy} & \cellcolor{darkgreen!10} 52.4\% & \textcolor{darkgreen}{+}3.1\% \\
  \end{tabular}
  \caption{Accuracy of ConvNeXt Femto on the Oxford Pets dataset with transferred filters from HorNet Tiny trained on the Food 101 dataset.}
  \label{tab:cross_domain_arch}

\end{table}

The results suggest that there are general spatial filter sets learned by depthwise convolutions, regardless of the architecture and dataset domain.

\section{Conclusions and Discussion}
This paper introduces the Master Key Filters Hypothesis, that there exist master key filter sets that are general, and the depthwise filters tend to converge to them. We provide evidence that DS-CNNs learn depthwise convolutional filters that remain general across diverse datasets, domains, and model architectures. 

Our experiments, spanning semantically divided ImageNet, cross-domain, and cross-architecture transfers, reveal that these filters maintain generality even in deeper layers, challenging prevailing notions that there is a transition from general to specialized filters, and filters get increasingly specialized in deeper layers of the network. 

When transferring the pointwise layers, we observed convergence issues across all datasets, even in selffer models, suggesting optimization difficulties. This may be attributed to the higher parameter count in pointwise layers, potential sparsity effects, and restricted permutation symmetries compared to depthwise layers. These findings align with prior work on fragile co-adaptation in neural networks~\cite{NIPS2014_375c7134}, where freezing certain layers can create challenging loss landscapes for training the remaining parameters. Future work could further investigate the specific mechanisms behind these optimization challenges.

The generality of depthwise filters has significant implications for transfer learning, enabling performance improvements when transferring filters from larger to smaller datasets, regardless of domain differences. Our results also open new avenues for cross-architecture knowledge transfer. But more important than all, these findings contribute to our understanding of the fundamentals of convolutional neural networks.

%Future work could explore understanding the interplay between spatial and channel-wise representations, examining how each contributes to the network's overall performance and adaptability. 
%Furthermore, the generality and transferability of spatial features in DS-CNNs present a promising opportunity for developing efficient transfer learning and domain adaptation techniques.

\section{Acknowledgments} Zahra Babaiee and Radu Grosu are supported by the
Austrian Science Fund (FWF) project MATTO-GBM I 6605. Peyman M. Kiasari is supported by
the TU Wien TrustACPS PhD School program that is supported by TTTech Auto and
B\&C Privatstiftung.

%In this paper, we introduced the Master Key Filters Hypothesis and presented compelling evidence for demonstrating that DS-CNNs learn depthwise convolutional filters that remain general across diverse datasets, domains, and model architectures. Our experiments, ranging from semantically divided ImageNet to cross-domain and cross-architecture transfers, reveal that these filters maintain their generality even in deeper layers, contrasting with prevailing notions from traditional CNNs. The generality of these filters has significant implications for transfer learning, enabling performance improvements when transferring filters from larger to smaller datasets, regardless of domain differences. Furthermore, our results open new avenues for cross-architecture knowledge transfer.

\bibliography{aaai25}

\end{document}